\documentclass[letterpaper, 10 pt, conference]{ieeeconf}
\usepackage[utf8]{inputenc}
\usepackage{cite}

  \usepackage[dvips]{graphicx}
  \usepackage{color}
  \usepackage{psfrag}

  \graphicspath{ {fig/} } 
\usepackage{amsmath}
\usepackage{amssymb}
\usepackage{bm}
\usepackage{algpseudocode}
  \usepackage[caption=false,font=footnotesize]{subfig}
\usepackage[hidelinks]{hyperref}

\usepackage{changes}

\usepackage[]{algorithm2e}

\usepackage{arydshln}

\newcommand*{\etal}{\emph{et al.}\@\xspace}
\newcommand*{\refsec}[1]{Section \ref{sec:#1}}
\newcommand*{\refig}[1]{Fig. \ref{fig:#1}}
\newcommand*{\refeq}[1]{Eq. (\ref{eq:#1})}

\let\oldsubsubsection\subsubsection
\renewcommand{\subsubsection}[1]{\vspace{2mm}\oldsubsubsection{#1}}

\graphicspath{ {fig/} } 
\renewcommand*{\vec}[1]{\bm{#1}}
\newcommand*{\set}[1]{\mathcal{#1}}
\usepackage{bm}

\IEEEoverridecommandlockouts 
\title{\LARGE \bf
Improved Reinforcement Learning Pushing Policies via Heuristic Rules
}
\author{%
 Marios Kiatos$^{1}$, Iason Sarantopoulos$^{1}$, Sotiris Malassiotis$^{2}$ and Zoe Doulgeri$^{1}$
\thanks{The research leading to these results has received funding from the European Community’s Framework Programme Horizon 2020 under grant agreement No 871704, project BACCHUS.}
\thanks{The first two authors have contributed equally.}
\thanks{$^1$Department of Electrical and Computer Engineering, Aristotle University of Thessaloniki, Thessaloniki, 54124 Greece {\tt\small \{mkiatos, iasons, doulgeri\}@ece.auth.gr}}%
\thanks{$^2$Information Technologies Institute (ITI) Center of Research and Technology Hellas (CERTH) 57001 Thessaloniki, Greece {\tt\small malasiot@iti.gr}}%
}

\begin{document}

\maketitle

\begin{abstract}
Non-prehensile pushing actions have the potential to singulate a target object from its surrounding clutter in order to facilitate the robotic grasping of the target. To address this problem we utilize a heuristic rule that moves the target object towards the workspace's empty space and demonstrate that this simple heuristic rule achieves singulation. We incorporate this effective heuristic rule to the reward in order to train more efficiently reinforcement learning (RL) agents for singulation. Simulation experiments demonstrate that this insight increases performance. Finally, our results show that the RL-based policy implicitly learns something similar to one of the used heuristics in terms of decision making. Qualitative results, code, pre-trained models and simulation environments are available at \url{https://github.com/robot-clutter/improved_rl}.
\end{abstract}

\section{Introduction}
Despite the research effort during the last decades, extracting target objects from unstructured cluttered environments via robotic grasping is yet to be achieved. Until to this day, robotic solutions are mostly limited to highly structured and predictable setups within industrial environments. In these environments the robots perform successfully because the tasks require characteristics such as precision, repeatability and high payload. However, tasks such as tight packing \cite{wang21}, throwing objects on target packages \cite{zeng20} or object grasping and manipulation in cluttered environments \cite{zeng19} require a much more different skill set and are still performed mostly by humans. Even though humans lack precision and strength, they are equipped with dexterity and manipulation skills that are missing from today's robots, endowing humans with the ability to deal effectively with extremely diverse environments. Such a fundamental manipulation skill is robust grasping of target objects \cite{bohg14, sarantopoulos18a, kiatos20}. However, when dealing with clutter, grasping affordances are not always available unless the robot intentionally rearrange the objects in the scene. As a result, non-prehensile manipulation primitives are used for object rearrangement \cite{stuber20} and a body of research emerged for total or partial singulation of objects using pushing actions, in order to create space around the target object required by the robotic fingers for grasping it. Total singulation \cite{sarantopoulos2021total} refers to the creation of free space around the object, while partial singulation refers to the creation of free space along some object sides.

Both hand-crafted and data-driven singulation methods have been developed for object singulation. Hand-crafted methods include analytic or heuristic methods, where analytic use exact models for making predictions assuming accurate knowledge of the environment and heuristic methods follow simple rules that can lead to task completion without assuming accurate knowledge of the environment. Analytic or heuristic-based methods use insights for the task, lacking however the generalization ability of the data-driven methods. The success of deep learning \cite{lecun15} in fields like computer vision or language processing motivated the robotics community to solve robotic problems using deep learning \cite{sunderhauf18, kroemer21} or reinforcement learning \cite{kober13, polydoros17, ibartz21, levine17}. However, deep learning approaches come with the cost of large data requirements, a problem that is exacerbated in robotics due to huge state/actions spaces as well as time consuming, costly and risky training when dealing with real robots. This led to training agents in simulated environments and transfer them in the real robotic system either by robust feature selection \cite{sarantopoulos2021total} or by sim-to-real methods \cite{rusu17}. 

In this paper, our objective is to study the learning-based policy proposed in our previous work \cite{sarantopoulos2021total} with respect to pushing policies that follow simple but effective heuristic rules, answering two questions:
Can the heuristics that achieve total singulation be used during training to increase the performance of the RL-based policy? Does the RL-based policy implicitly learn something similar to the heuristics?  In this work, we aim to answer the above questions by:
\begin{itemize}
    \item demonstrating that heuristic policies that move the target object towards the empty-from-obstacles space can be effective in a singulation task,
    \item demonstrating that guiding the training with these heuristic rules in the reward, increase the performance of the RL-based policy and
    \item performing experiments to study the similarity of the initial RL-based policy with the heuristic-based policies, in terms of decision making. 
\end{itemize}

In the following section the related work is presented. \refsec{problem_formulation} describes the environment and the problem formulation. \refsec{pushing_policies} presents the heuristic and the learning based policies used in this work as well as training details for the learning-based policies. \refsec{results} provides the results from our experiments and finally \refsec{conclusions} draws the conclusions.

\section{Related work}
\subsection{Analytic and heuristic-based approaches}
Lynch \etal \cite{lynch96} pioneered research on analytic models of push mechanics. Following works focused on planning methods to reduce grasp uncertainty in cluttered scenes by utilizing push-grasping actions \cite{dogar11, dogar12}. Cosgun \etal \cite{cosgun11} proposed a method for selecting pushing actions in order to move a target object in a planned position as well as clearing the obstructed space from other objects, assuming however knowledge of exact object poses. In contrast to these analytic methods where exact knowledge of the objects and the environment is assumed, other works employ heuristics for manipulating table-top objects. Hermans \etal \cite{hermans12} proposed a method that separates table-top objects, exploiting the visible boundaries that the edges of the objects create on the RGB image. Similarly, Katz \etal \cite{katz13} segmented the scene for poking the objects in the scene before grasping them. Chang \etal \cite{chang12} proposed a heuristic singulation method which pushes clusters of objects away from other clusters. Their method checked if a push separated the target cluster into multiple units and iteratively kept pushing until the target cluster remained a single unit after the push. The downside is that the clusters should be tracked in order to produce future decisions. Finally, Danielczuk \etal \cite{danielczuk18} proposed two heuristic push policies for moving objects towards free space or for diffusing clusters of objects, using the euclidean clustering method for segmenting the scene into potential objects. We adapt the free space heuristic to our problem and demonstrate that achieves total singulation.

\subsection{Learning-based approaches}
Recently, a lot of researchers employed learning algorithms for object manipulation. Boularias \etal \cite{boularias2015learning} explored the use of reinforcement learning for training policies to choose among grasp and push primitives. They learnt to singulate two specific objects given a depth image of the scene but the agent required retraining for a new set of objects. Zeng \etal \cite{zeng18a} used $Q$-learning to train end-to-end two fully convolutional networks for learning synergies between pushing and grasping, leading to higher grasp success rates. Yang \etal \cite{yang20} adapted \cite{zeng18a} to grasp a target object from clutter by partially singulating it via pushing actions. Kurenkov \etal \cite{kurenkov20} and Novkovic \etal \cite{novkovic20} used deep reinforcement learning to learn push actions in order to unveil a fully occluded target object. In contrast to singulation, achieving full visibility of the target does not always create free space between the target and surrounding obstacles, which is necessary for prehensile grasping. On the contrary to model-free reinforcement learning approaches, Huang \etal \cite{9591286} proposed a model-based solution for target object retrieval. Specifically, they employ visual foresight trees to intelligently rearrange the surrounding clutter by pushing actions so that the target object can be grasped easily. All the above approaches focus on the partial singulation of objects in order to grasp them.

On the other hand, many works develop methods that totally singulate the objects. Eitel \etal \cite{eitel2020learning} trained a convolution neural network on segmented images in order to singulate every object in scene. However, they evaluated a large number of potential pushes to find the optimal action.  Kiatos \etal \cite{kiatos19} trained a deep $Q$-network to select push actions in order to singulate a target object from its surrounding clutter with the minimum number of pushes using depth features to approximate the topography of the scene. Sarantopoulos and Kiatos \cite{sarantopoulos20} extended the work of \cite{kiatos19} by modelling the $Q$-function with two different networks, one for each primitive action, leading to higher success rates and faster network convergence.  Finally, Sarantopoulos \etal \cite{sarantopoulos2021total} proposed a modular reinforcement learning approach which uses continuous actions to totally singulate the target object. They used a high level policy, which selects between pushing primitives that are learnt separately and incorporated prior knowledge, through action primitives and feature selection, to increase sample efficiency. Their experiments demonstrated increased success rates over previous works. We build upon this work to demonstrate that incorporating insights from heuristics to reinforcement learning algorithms via reward shaping, can increase the performance further and study the similarity of the decisions that this learning-based policy makes with respect to the heuristic-based policies.

\begin{figure}
  \begin{center}
  \includegraphics[width=0.4\textwidth]{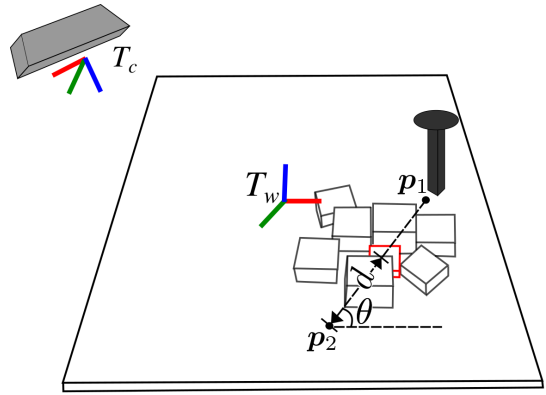}
  \end{center}
  \caption{Illustration of the environment including the support surface, the objects, the robotic finger and the camera. The inertia frame $T_w$ is placed on the center of the table. A push action is defined by the initial position $\bm{p}_1$ that the push starts, the direction angle $\theta$ and the pushing distance $d$. The goal is to singulate the target object (red contour) from the surrounding obstacles.}
  \label{fig:environment}
\end{figure}

\section{Problem formulation}\label{sec:problem_formulation}
The objective is to properly rearrange the clutter in order to totally singulate the target object. Total singulation means that the distance between the target object and its closest obstacle is larger than a predefined constant distance $d_{sing}$. Given an RGB-D image of the scene, the agent selects the pushing action that maximizes the probability of singulating the target object.

\begin{figure}[ht!]
  \begin{center}
  \includegraphics[width=0.45\textwidth]{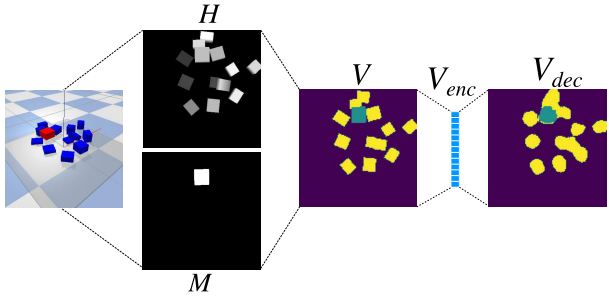}
  \end{center}
  \caption{Illustration of the visual state generation $V$, representing a segmentation of the scene. $H$ denotes the heightmap of the scene and $M$ the mask of the target object. An autoencoder is used for dimensionality reduction of $V$ to the encoding vector $V_{enc}$. $V_{dec}$ is the reconstructed visual representation by the decoder.}
  \label{fig:state}
\end{figure}

\subsection{Environment}
The environment consists of a robotic finger, a single overhead RGB-D camera and rigid objects in clutter resting on a rectangular workspace with predefined dimensions $s \times s$ and adequate surrounding free space. The pose, dimensions and the number of obstacles are random. Finally, the world frame $\{T_w\}$ is placed to the center of the workspace with its z-axis being opposite to the gravity direction as shown in \refig{environment}.

\subsection{Visual Representation}\label{sec:visual_representation}
Similar to \cite{sarantopoulos2021total}, we use a visual representation $V$, which is extracted from the camera observation and is generated by the fusion of the 2D heightmap $H$ and the mask $M$ (\refig{state}). Specifically, the heightmap $H$ shows the height of each pixel and is generated from the depth image by subtracting from each pixel the maximum depth value. The mask $M$ shows which pixels belong to the target object and is extracted by detecting the visible surfaces of the target object. Note that we assume that the target is at least partially visible and can be visually distinguished from the obstacles. We also crop the heightmap and the mask to keep only information within the workspace. Essentially, the visual state representation is an image that provides a segmentation of the scene between the target, the obstacles and the support surface. The pixels that belong to the target have the value 0.5, those belonging to an obstacle the value 1 and the rest of them belonging to the support surface the value 0. Finally, we resize the visual representation to 128 $\times$ 128. An illustration of the visual state representation is shown in \refig{state}.

\subsection{Actions}\label{sec:actions}
A push action is defined by a line segment $[\vec{p}_1, \vec{p}_2] \in \mathbb{R}^2$, which is parallel to the workspace and is placed at a constant height $h$ above it. In contrast to previous work \cite{sarantopoulos2021total}, we use only the push-target primitive, which always pushes the target object and is defined by a vector $[d, \theta]$ where $d$ is the pushing distance and $\theta$ is the angle defining the direction of push. The height $h$ is selected by the primitive at the half space between the table and the top of the target. This restricts the available pushing positions since the surrounding obstacles may be taller than this height. To ensure that the finger will avoid undesired collisions with obstacles during reaching the initial position we define the initial position $\vec{p}_1$ to be the closest to the target position which is not occupied by obstacles along the direction $\theta$. Note, however, that during the execution of the pushing action, interaction with the obstacles may occur but not considered undesired. To achieve this obstacle avoidance, we use the visual representation $V$ and find the first patch of pixels along direction defined by $\theta$ that contains only pixels from the table. Given the vector $[d, \theta]$, the final point $\vec{p}_2$ is computed as $\vec{p}_2 = \vec{p}_1 + d[\cos(\theta), \sin(\theta)]^T$. As a result, all the policies will try to move only the target object. Of course during the execution of the primitive, some obstacle is possible to move. 

\section{Pushing Policies}\label{sec:pushing_policies}
In this section we will present the different pushing policies used in this work. We use two types of policies: heuristic policies that take greedy decisions based on a heuristic rule and learned policies which are optimized using deep reinforcement learning.

\subsection{Heuristic-based pushing policies}\label{sec:heuristics}
\begin{figure*}[!t]
  \begin{center}
  \includegraphics[width=0.9\textwidth]{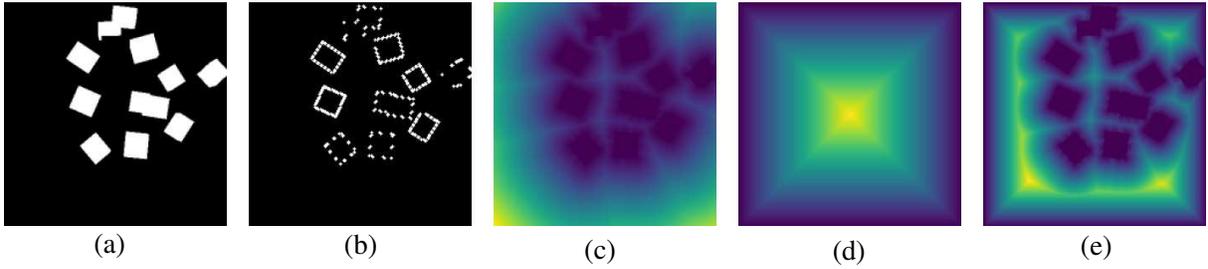}
  \end{center}
  \caption{The different stages for extracting the empty space map ESM. Purple color denotes low distance values and yellow color high distance value. (a) The initial mask of the obstacles. (b) The points belonging to the contour $\set{C}$ of the obstacles' mask. (c) The obstacles distance map ODM showing the distances from the obstacles'. (d) The map limits distance map LDM showing the distances from the workspace limits. (e) The final empty space map ESM which combines ODM and LDM.}
  \label{fig:empty_space_map}
\end{figure*}
The heuristic policy used in this work is a policy which selects pushing actions for moving the target object towards the empty space of the scene, which is inspired by \cite{danielczuk18}. The assumption here is that moving the target object towards the empty space will create empty space around it, and thus accomplishing total singulation.

In order to select the best action, we need to calculate the empty space map (ESM) shown in \refig{empty_space_map}(e). To produce this map, we first create a mask of the obstacles of the scene (\refig{empty_space_map}(a)) by removing the target object from the visual state representation $S$, since we only interested in calculating the distances from the obstacles. Then, we calculate the points $\vec{c}\in C$ that belong to the contour $C$ of this mask (\refig{empty_space_map}(b)). Subsequently, we create the obstacle distance transform (ODT) showing the distances of each workspace point from the obstacles (\refig{empty_space_map}(c)). Hence, the value of each pixel $(i, j)$ of ODT is its minimum distance from every point of the contour $\mathcal{C}$, i.e.: 
\begin{equation}
\text{ODT}(i, j) = \min_{\vec{c}\in C}|[i, j]^T - \vec{c}|
\end{equation}
To avoid pushing the target object out of the workspace limits, we create a limits distance map (LDM) by calculating for each pixel its minimum distance from the limits of the workspace, as shown in \refig{empty_space_map}(d). Then, the final map is given by keeping the minimum value between the obstacles distance map and the limit distance map: 
\begin{equation}
\text{ESM}(i, j) = \min\{\text{ODT}(i, j), \text{LDM}(i, j)\}
\end{equation}
as shown in \refig{empty_space_map}(e). As a final step we normalize the values of the map to $[0, 1]$, given its minimum and maximum value.

\begin{figure}[!t]
  \begin{center}
  \includegraphics[width=0.48\textwidth]{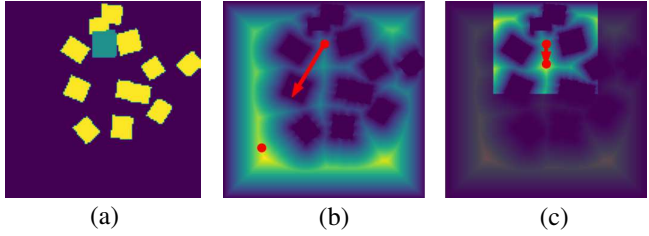}
  \end{center}
  \caption{Comparison of decision making between the heuristic policies. (a) The visual state representation of the scene. (b) The ES policy. (c) The LES policy.}
  \label{fig:global_vs_local}
  \vspace{-0.6cm}
\end{figure}

Given the empty space map, we can select the direction $\theta$ and the pushing distance $d$ for pushing the target object towards the empty space. In order to determine the direction, we find all the pixels $\vec{p}$ of ESM map whose value is above 0.9, i.e. the top 10\%. From this set, we select as the optimal pixel $\vec{p}^*$ the pixel that has the minimum distance from the centroid  $\vec{o}$ of the target object, calculated using the target's mask $M$. Note that taking multiple best values, instead of just the maximum, can produce better decisions. For example, if one pixel far from the target object has the value 0.97 and another pixel much closer to the target has the value 0.95, it would not be efficient to just greedily select the maximum value, because the agent could oscillate between pushes that unnecessarily attempt to reach remote areas. Then, the direction $\theta$ is computed by the line segment between $\vec{o}$ and $\vec{p}^*$ and the pushing distance $d$ is the length of this line segment. Finally, we enforce an upper limit to $d$ for avoiding long pushes that can result to unpredictable behaviour of the target object.  \refig{global_vs_local}(a) shows the selected push of this policy, called Empty Space Policy (ES) for this scene.

Even though ES policy selects the optimal pixel between multiple candidates, it can still produce pushes based on regions that are far from the target. For this reason, we also propose a variant called Local Empty Space policy (LES) that selects the best action taking into account only a local neighborhood of the target object. To this end, we crop the ESM map around the target object, producing a local empty space map (LESM) shown in \refig{global_vs_local}(b) along with the selected action by LES policy. In particular, the crop size corresponds to 1/4 of the workspace area. Notice that for the same scene, the two policies, ES and LES, produce very different actions.

\subsection{Learning-based pushing policies}\label{sec:learning_policies}
To train the reinforcement learning policies we formulate the problem as a Markov Decision Process (MDP) with time horizon $T$. An MDP is a tuple ($\mathcal{S}, \mathcal{A}, \mathcal{R}, \mathcal{T}, \gamma$) of continuous states $\mathcal{S}$, continuous actions $\mathcal{A}$, rewards $\mathcal{R}$, an unknown transition function $\mathcal{T}$ and a discount factor $\gamma$. 

\textbf{\textit{States:}}
We use both the visual state representation presented in \refsec{visual_representation} and the full state representation taken from simulation. The full state representation contains the exact poses of the objects as well as their bounding boxes. To reduce the dimensionality of the visual state representation, we train an autoencoder on a large dataset of observations $V$ and use the latent vector $V_{enc}$ as the input to the agents (\refig{state}). The architecture of the autoencoder is similar to the one proposed in \cite{sarantopoulos2021total}, but trained with cross-entropy loss instead of regression. Note that in this work we encode the whole view of the scene instead of a downscaled crop centered around the target.

\textbf{\textit{Actions:}}
All the agents use the pushing actions defined in \refsec{actions}.

\textbf{\textit{Rewards:}}
First we define a sparse reward that will motivate the RL agent to discover a policy purely for the singulation task. Then, to investigate how insights from the heuristics, described in \refsec{heuristics}, improve the performance of reinforcement learning policies, we present two additional reward schemes. To achieve that, we use the ESM and LESM maps for reward shaping. Specifically, we measure the error $e$ between the predicted angle $\theta_{p}$ that the agent produced and the angle $\theta_{h}$ that the heuristic policy would have taken if used:
\begin{equation}\label{eq:error}
    e(\theta_p, \theta_h) = \frac{1}{2} \left(1 - \cos(|\theta_h - \theta_p|)\right) \in [0, 1],
\end{equation}
This error will be zero if the directions are the same ($|\theta_h -\theta_p| = 0$) and it will be one if the directions are exactly opposite ($|\theta_h -\theta_p| = \pi$). To this end, we propose three different versions of reward:
\begin{enumerate}
    \item For training the vanilla RL agent, we assign $r=-1$ if the target falls out of the workspace and $r=+1$ if the target is totally singulated. In any other case, we assign a small negative reward $r=-1/T_{max}$ to each push for motivating the agent to minimize the number of actions, where $T_{max}$ is the maximum number of timesteps.
    \item To guide the RL policy towards the global empty space, we use the ESM map to compute the error $e$. Specifically, we assign $r = (-1 - e) / (2T_{max})$ for each push in order to penalize the pushes that move the target away from the optimal $\vec{p}^*$ computed from the ESM map and simultaneously minimize the number of actions. Moreover, we assign $r=-1$ if the target falls out of the workspace and $r=+1$ if the target is totally singulated.
    \item Finally, to guide the RL policy towards the local empty space, we use the same reward as above but utilize the LSM map to compute the error $e$.
\end{enumerate} 

\textbf{\textit{Terminal states:}}
The episode is terminated if the target is totally singulated, falls out of the workspace or the maximum number of timesteps has been reached. 

We exploit the fact that we train in simulation by employing an asymmetric actor-critic algorithm, proposed in \cite{pinto18}, in which the critic is trained with the full state representation while the actor gets the visual state representation as inputs. In particular, we use the Deep Deterministic Policy Gradient (DDPG) \cite{lillicrap16} as the actor-critic algorithm. In contrast to \cite{sarantopoulos2021total} we train the agents in an off-policy way by filling a replay buffer with 100K transitions of random pushing actions. During training we do not update the replay buffer with new transitions and thus we can pass the entire dataset multiple times (epochs) to update the networks.

The actor and the critic are both 3-layered fully-connected with 512 units each layer. The target networks are updated with polyak averaging of 0.999. We train the networks for 50 epochs with batch size 32 and discount factor $\gamma = 0.9$. Both networks are trained using Adam optimizer with learning rate $10^{-3}$. 

\section{Results}\label{sec:results}
The objective of the experiments are two-fold: 
(a) to evaluate if the use of heuristics during training can improve the performance of the RL-based policy in the singulation task and (b) to examine if there is any similarity in the decision making between the policy produced by reinforcement learning and the heuristics. The policies that we evaluate are the following:

\begin{itemize}
    \item ES: a heuristic policy that chooses pushing actions according to the empty space map of \refsec{heuristics}.
    \item LES: a heuristic policy which selects the optimal action based on the local empty space map of \refsec{heuristics}.
    \item RL: policy trained with the reward (1) of \refsec{learning_policies}.
    \item RL-ES: policy trained with the reward (2) of \refsec{learning_policies}, that guides the agent using the empty space map.
    \item RL-LES: policy trained using the reward (3) of \refsec{learning_policies} that guides the agent using the local empty space map.
\end{itemize}
We use the Bullet physics engine \cite{coumans2016pybullet} to advance simulation. The objects are approximated as rectangulars of random dimension.
The number of obstacles is between 8 and 13.
Furthermore, we use a rectangular robotic finger, $d_{sing}=3$ cm, a workspace $0.5 \times 0.5$ m and keep all the dynamic parameters of the simulation fixed. All polices are evaluated in the same 200 scenes in simulation.

\subsection{Performance of the policies}
The proposed heuristics where developed to solve the singulation task in an environment with adequate empty space. As we expected ES and LES achieve high success rates and in particular 97.9\% and 98.8\% respectively. These results show the effectiveness of these heuristics and consequently they can be used as insights to the RL-based policies for improving their performance.

\begin{figure}[t!]
  \begin{center}
  \includegraphics[width=0.45\textwidth]{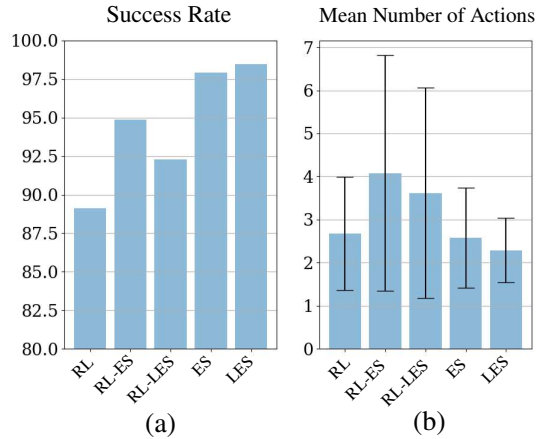}
  \end{center}
  \caption{Qualitative results for the performance of the different policies. (a) Success rates; (b) Mean number of actions}
  \label{fig:success}
\end{figure}

As we see in \refig{success}(a), RL-ES and RL-LES policies outperform RL policy. The improved success rates demonstrate the importance of incorporating insights from heuristics to guide reinforcement learning agents through reward shaping. This leads to faster convergence and improved success rates. Finally, the RL-ES policy outperforms the RL-LES due to the fact that it is optimized with ES map which takes into consideration the whole scene and thus producing actions that lead to increased success rate.


\begin{table}[!t]
  \begin{center}
    \caption{Results}
    \label{tab:results}
    \begin{tabular}{l | c c c}
      \hline
      \textbf{Policy} & \textbf{Success} & \textbf{Mean} & \textbf{Std}  \\
         & \textbf{rate \%} & \textbf{\#actions} & \textbf{\#actions}  \\
      \hline
      RL  &  89.1 & \textbf{2.67} & \textbf{1.31}\\
      RL-ES &  \textbf{94.8} & 4.08 & 2.73\\
      RL-LES  & 92.3  & 3.67 & 2.39\\
      \hline
    \end{tabular}\\[5pt]
    ES = Empty Space, LES = Local Empty Space
  \end{center}
\end{table}

\begin{figure*}
  \begin{center}
  \includegraphics[width=0.7\textwidth]{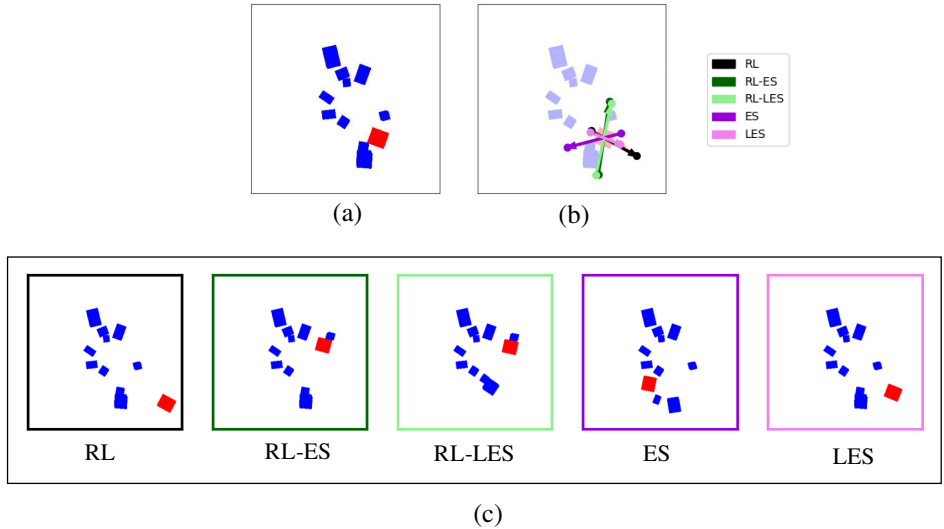}
  \end{center}
  \caption{A sample of actions taken by the policies in the same scene. (a) The initial state of the scene. (b) The actions taken by each policy. The start and the end of the arrow is placed on the initial and the final position of the push. (c) The next states for each action.}
  \label{fig:transitions}
\end{figure*}

\begin{figure}[t!]
  \begin{center}
  \includegraphics[width=0.5\textwidth]{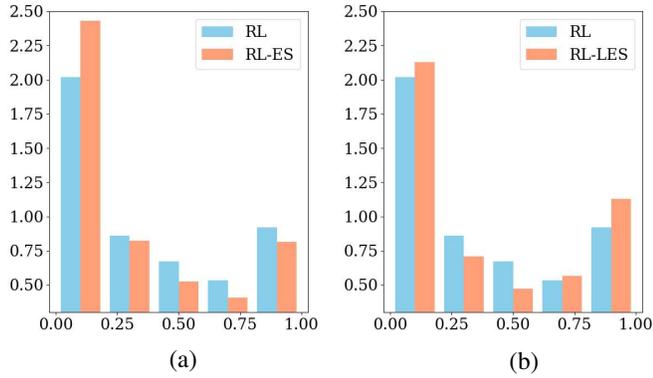}
  \end{center}
  \caption{The distribution of the errors between the directions chosen by the policies. (a) The errors between the RL, RL-ES policies with the heuristic ES. With blue color the $e(\theta_{RL}, \theta_{ES})$ and with orange color the error $e(\theta_{RL-ES}, \theta_{ES})$;  (b) The errors between the RL, RL-LES policies with the heuristic LES. With blue color the $e(\theta_{RL}, \theta_{LES})$ and with orange color the error $e(\theta_{RL-LES}, \theta_{LES})$.}
  \label{fig:error_distribution}
\end{figure}

\subsection{Similarity of decision making between learning and heuristics}
For evaluating the similarity of the decision making between heuristic and learning based policies, we compare the pushing directions decided by the policies RL, RL-ES, RL-LES with the pushing directions that would have been decided by the heuristic policies ES, LES if used in the same scenes. We only consider the differences in the direction and not in the pushing distance as the latter is dependant in the direction. To this end, we run 200 episodes for each policy RL, RL-ES, RL-LES. For each scene we compute the angle of the pushing direction $\theta_p$ produced by each policy and executed in the environment and the angle of the pushing direction $\theta_h$ that the heuristic ES or LES would have produced, if used. Then, we calculate the error between these angles $e(\theta_p, \theta_h)$ from \refeq{error}. 

\refig{error_distribution}(a) shows the distribution of the errors for the 200 episodes between RL and ES policies (blue bars) along with the distribution of errors between RL-ES and ES policies (orange bars). Notice that the distribution of RL-ES is shifted towards smaller errors compared to the RL distribution (see the difference in the range of 0-0.2 of the error), indicating that RL and RL-ES decide on different actions for the same scene and thus RL learns something different. On the other hand, \refig{error_distribution}(b) shows the distribution of the errors between RL and LES policies (blue bars) along with the distribution of errors between RL-LES and LES policies (orange bars). In this figure we observe that the distribution of errors between RL and RL-LES are matching, indicating similar decision making. From the above, we can conclude that the RL policy implicitly learns something more similar to the local heuristic than to the global heuristic. \refig{transitions} demonstrates a sample of actions taken by the different policies. Notice that in this sample the RL policy decide on a similar action with the local policies LES and RL-LES than the global ES and RL-ES.

This conclusion can also explain the results of \refig{success}(a) related to the performance of the policies. If the RL agent learns implicitly something similar to the LES policy, then we expect that guiding with directions from LES should have a smaller impact in the performance than guiding with ES. As we see in \refig{success}(a), guiding the RL agent to learn a policy closer to ES, the success rate was increased from 89.1\% (RL) to 94.8\% (RL-ES), in contrast to guiding the agent to learn a policy closer to LES, where the corresponding improvement of the success rate was smaller (92.3\% for RL-LES).

Finally, the conclusion that RL learns something closer to the local heuristic LES is to be expected from another point of view. The discount factor $\gamma$ limits the time horizon that the agent can optimize the Q-values into the future and given the limited pushing distance in this task, the agent learns to act better by taking into account observations from the local neighborhood of the target object, rather than observations corresponding to a distance far from the target.

\section{Conclusions}\label{sec:conclusions}
In this paper, we employed simple heuristic rules to achieve total singulation by pushing the target object towards the free space of the workspace. Experiments demonstrated that if we incorporate these heuristics to the reward of RL agents, singulation success is increased. Finally, we showed that the pure RL-based policy implicitly learns something more similar to the local heuristic.

\bibliographystyle{IEEEtran}
\bibliography{references.bib}

\end{document}